\documentclass{article}

\usepackage{arxiv}

\usepackage[utf8]{inputenc} 
\usepackage[T1]{fontenc}    
\usepackage{booktabs}       
\usepackage{amsfonts}       
\usepackage{nicefrac}       
\usepackage{microtype}      
\usepackage{lipsum}		

\usepackage{doi}

\usepackage{graphicx}

\usepackage{listings}
\usepackage{cite}
\usepackage{amsmath,amssymb,amsfonts}
\usepackage{algorithmic}
\usepackage{graphicx}
\usepackage{textcomp}
\usepackage{xcolor}
\usepackage{placeins}
\usepackage{breqn}
\usepackage{makecell}
\usepackage{booktabs}
\usepackage{listings}
\usepackage{subcaption}
\usepackage{url}
\usepackage{hyperref}
\usepackage[multiple]{footmisc}
\usepackage{flushend}
\usepackage{adjustbox}  
\usepackage[inline]{enumitem}

\usepackage[font=small,skip=3pt]{caption}
\usepackage[subrefformat=parens,labelformat=parens]{subcaption}
\title{Deep Learning-Based Intrusion Detection for Automotive Ethernet: Evaluating \& Optimizing Fast Inference Techniques for Deployment on Low-Cost Platform\thanks{    This is a preprint of an article that has been accepted for publication in the proceedings of SBC/BRACIS 2025 and the final version will be published by Springer in \textit{Lecture Notes in Computer Science (LNCS)}. The final published version may contain differences.}}

\author{%
    \textbf{Pedro R. X. Carmo} \\
    Centro de Informática (CIn) \\
    Grupo de Pesquisa em Redes e Telecomunicações\\(GPRT)\\
    Universidade Federal de Pernambuco (UFPE) \\
    Recife, Brasil \\
    \texttt{prxc@cin.ufpe.br}
    \and
    \textbf{Igor de Moura} \\
    Centro de Informática (CIn) \\
    Universidade Federal de Pernambuco (UFPE) \\
    Recife, Brasil \\
    \texttt{imp2@cin.ufpe.br}
    \and
    \\
    \textbf{Assis T. de Oliveira Filho} \\
    Centro de Informática (CIn) \\
    Grupo de Pesquisa em Redes e Telecomunicações\\(GPRT)\\
    Universidade Federal de Pernambuco (UFPE) \\
    Recife, Brasil \\
    \texttt{atof@cin.ufpe.br}
    \and
     \\
    \textbf{Djamel Sadok} \\
    Centro de Informática (CIn) \\
    Grupo de Pesquisa em Redes e Telecomunicações\\(GPRT)\\
    Universidade Federal de Pernambuco (UFPE) \\
    Recife, Brasil \\
    \texttt{jamel@cin.ufpe.br}
    \and
     \\
    \textbf{Cleber Zanchettin} \\
    Centro de Informática (CIn) \\
    Universidade Federal de Pernambuco (UFPE) \\
    Recife, Brasil \\
    \texttt{cz@cin.ufpe.br}
}

\renewcommand{\shorttitle}{}

\hypersetup{
pdftitle={Deep Learning-Based Intrusion Detection for Automotive Ethernet: Evaluating \& Optimizing Fast Inference Techniques for Deployment on Low-Cost Platform},
pdfsubject={cs.LG, cs.NI},
pdfauthor={Pedro R.X.~Carmo}
}

\begin{document}

\maketitle

\begin{abstract}
Modern vehicles are increasingly connected, and in this context, automotive Ethernet is one of the technologies that promise to provide the necessary infrastructure for intra-vehicle communication. However, these systems are subject to attacks that can compromise safety, including flow injection attacks. Deep Learning-based Intrusion Detection Systems (IDS) are often designed to combat this problem, but they require expensive hardware to run in real time. In this work, we propose to evaluate and apply fast neural network inference techniques like Distilling and Prunning for deploying IDS models on low-cost platforms in real time. The results show that these techniques can achieve intrusion detection times of up to 727 $\mu$s using a Raspberry Pi 4, with AUCROC values of 0.9890.
\end{abstract}


\section{Introduction}

Connected Autonomous Vehicles (CAVs) rely heavily on connectivity features, including cameras, video-on-demand, and signal recognition. Vehicles are becoming more connected, and the demand for bandwidth is increasing. In this context came the Automotive Ethernet, which promotes speeds of up to 100 Mb/s. The IEEE 100BASE-T1 standard allows Ethernet to be deployed with a twisted-pair cable \cite{porter2018100base}. In addition to this standard, others were proposed to guarantee low latency, time sensitivity, and communication priority metrics. The AVB standard stands out, incorporating protocols like AVTP (Audio-Video Transport Protocol), essential for reliable, time-sensitive traffic \cite{matheus2021automotive}.
\par 
In the context of CAVs, it is important to consider that the feasibility and surface of malicious attacks on these systems are abundant. Various attacks can be performed on IVNs \cite{8053478}. For example, a DoS (Denial of Service) attack is simple to perform on automotive Ethernet. Once the attacker has access to the network, he only needs to flood false packets with high priority and prevent nodes from transmitting on the network. Intrusion Detection (IDSs) is the main method to be used as a countermeasure against cyber-physical attacks on IVNs \cite{8688625}.

In this work, we address the challenge of deploying Intrusion Detection Systems (IDS) for real-time protection in automotive Ethernet networks, particularly in the context of CAVs. While Deep Learning-based IDSs are effective, their high computational demands often require expensive hardware, limiting their practicality for in-vehicle use. To overcome this limitation, we propose to evaluate and apply fast neural network inference techniques, such as Distilling and Pruning, to optimize IDS models for deployment on low-cost platforms. {Although knowledge distillation and pruning have been extensively studied in general deep learning contexts, their systematic combination and evaluation for real‑time intrusion detection in automotive Ethernet on ultra‑low‑cost edge devices remains underexplored. The main \textbf{novel contributions} of this paper are:}
{
  \begin{enumerate*}[label=(\arabic*), itemjoin={{; }}, font=\bfseries]
    \item A \textbf{comprehensive comparison} of pruning, quantization, and distillation techniques applied to a 2D‑CNN IDS for automotive Ethernet
    \item The design and evaluation of an \textbf{ultra‑light student model} that meets real‑time thresholds ($\leq$1 ms) on Raspberry Pi 4 and Jetson Nano
    \item \textbf{Practical guidelines} for deploying optimized IDS on resource‑constrained embedded platforms.
  \end{enumerate*}}

To evaluate the feasibility of Deep Learning-based IDS in automotive networks, we use a real-world dataset based on the AVTP protocol \cite{b2} and explore fast inference techniques, such as pruning and knowledge distillation, to reduce computational complexity without compromising accuracy. The architecture used is based on a 2D-CNN model previously proposed in the literature \cite{b1}, which served as a reference for comparison with the optimized versions. The impact of these techniques was analyzed in terms of inference time and accuracy, with experiments conducted on low-cost hardware, such as the Raspberry Pi 4.

This study demonstrates the feasibility of these techniques for intrusion detection in Automotive Ethernet. While previous works focus on the effectiveness of Deep Learning-based IDS, our approach emphasizes optimizing these models for embedded devices, making their application more accessible and feasible for vehicular networks. In addition, we present a detailed analysis of the effects of pruning and distillation on IDS performance, providing practical guidelines for implementing efficient security systems in connected vehicles.


The rest of this paper is structured as follows: \textbf{Section 2} discusses related work. \textbf{Section 3} introduces the dataset used for training and evaluation. \textbf{Section 4} outlines our methodology. \textbf{Section 5} describes the experimental setup and evaluation process. \textbf{Section 6} presents the obtained results, followed by \textbf{Section 7}, which provides an in-depth discussion, analyzing trade-offs between detection accuracy, computational efficiency, and practical deployment. Finally, \textbf{Section 8} concludes the paper.

\section{Related Works and Background}

Intrusion detection in automotive Ethernet networks has been explored using deep learning-based models, particularly Convolutional Neural Networks (CNNs). Jeong et al. \cite{b1} proposed a CNN-based IDS for detecting malicious AVTP traffic, demonstrating high accuracy in identifying injected packets. Their work also introduced a dataset \cite{b2} containing AVTP packets labeled as benign or malicious, widely used for evaluating IDS models.

Knowledge distillation is widely used to optimize deep learning models for constrained platforms. Hinton et al. \cite{hinton2015distillingknowledgeneuralnetwork} introduced knowledge distillation, where a smaller student model learns from a larger, well-trained teacher model by minimizing a combination of standard cross-entropy loss and a distillation loss based on softened output probabilities. This approach enables compact models to retain high accuracy while significantly reducing inference time and computational costs.

Model compression methods like MobileViT \cite{b7} and MobileNet \cite{mobilenet}, emphasize efficient architectures for mobile and embedded devices. Additionally, pruning techniques \cite{tf_pruning_guide1, tf_pruning_guide2} and quantization methods \cite{tensorflowL2} have been applied to CNNs to further optimize deep learning models for real-time applications. In the context of automotive security, Carmo et al. \cite{b8} explored lightweight machine learning models for detecting cyberattacks in automotive Ethernet, reinforcing the necessity for IDS solutions that balance accuracy and computational efficiency.

Building on these works, this study applies knowledge distillation to optimize CNN-based IDS models for real-time execution on low-cost platforms. By leveraging distillation along with pruning and quantization, the proposed models promise substantial improvements in inference time while attempting to maintain high detection performance, making them suitable for deployment in resource-constrained automotive environments.

\section{Dataset}

\begin{figure}[!ht]
\centerline{\includegraphics[width=1\textwidth,keepaspectratio]{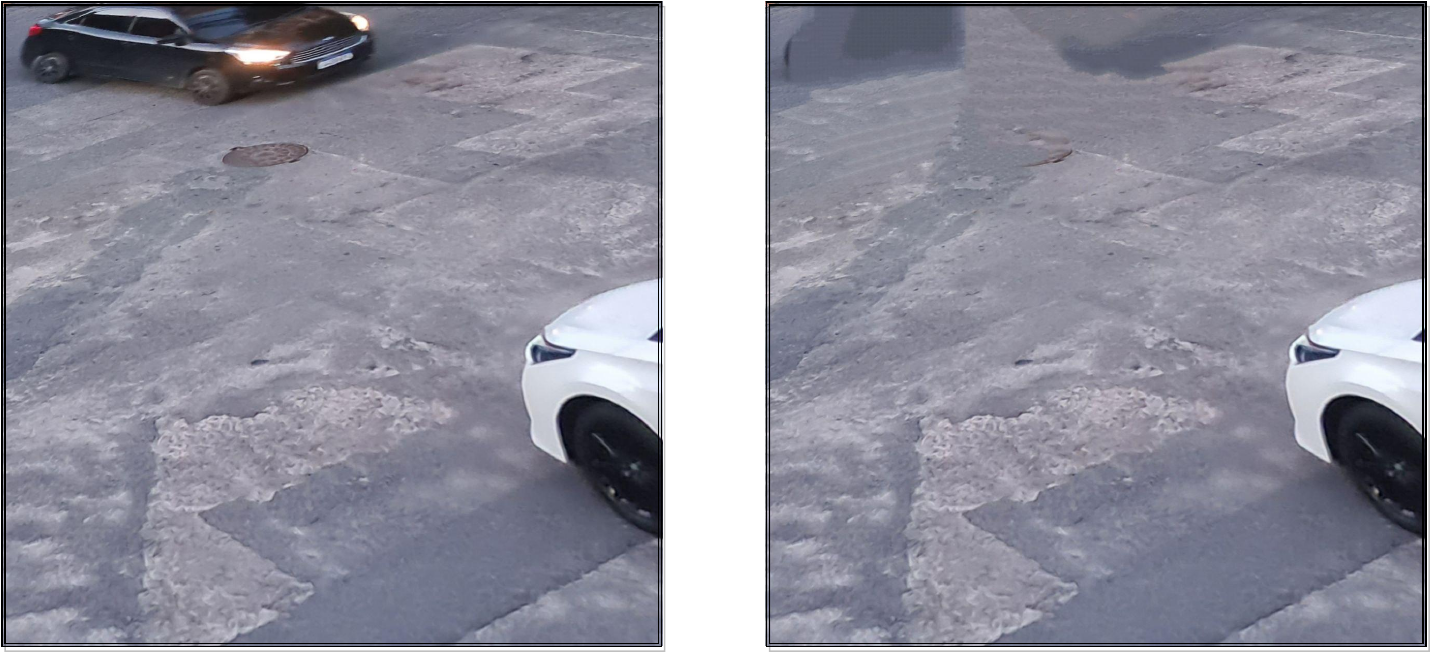}}
\caption{Real frame (left) and an injected frame (right). The injected frame was artificially created by replacing the original content with malicious data while preserving the visual structure. Reproduced from \cite{b8} with permission.}

\label{fig1}
\end{figure}

 The Dataset used in this work can be found in \cite{b2}. A real dataset in PCAP format. This dataset is made up of AVTP packets, where each packet is labeled as ``benign" (if it is a real packet) or ``injected/malicious" (if it is an injected packet by an attacker). The dataset creators collected attack-free AVTP-DU packets in lab and on-road scenarios. In addition, video frames were inserted into the dataset to simulate a \textit{replay attack}. In Fig \ref{fig1}, you can see an example of a real and injected frame.

To simulate a replay attack, the authors \cite{b1} send a video frame containing 36 AVTP-DU streaming packets while capturing the actual packets. This malicious frame is available in \cite{b2}, together with two datasets built on the data mentioned in the previous paragraph. In summary, the data set built inside the laboratory is called $D_{indoors}$, and the dataset built with the vehicle in motion is called $D_{driving}$.

\begin{figure}[!ht]
\centering
\includegraphics[width=1\textwidth,keepaspectratio]{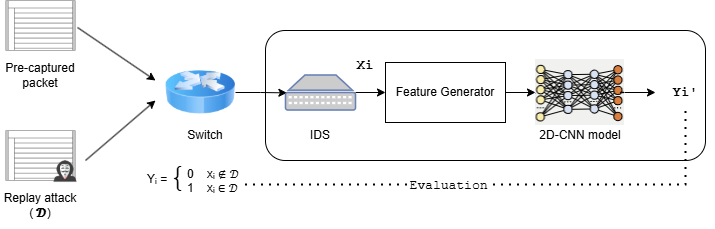}
\caption{Overview of system design. Pre-captured and replay attack packets are processed through the IDS, where features are extracted and classified using a 2D-CNN model. Adapted from \cite{b1}.}
\label{fig2}
\end{figure}

\section{Methodology}

    In this work, we evaluate the feasibility of deploying Deep Learning-based Intrusion Detection Systems (IDS) in Automotive Ethernet networks using techniques that optimize inference speed and resource efficiency. As a baseline, we replicated the 2D-CNN model proposed in \cite{b1} for intrusion detection in Automotive Ethernet, using the dataset from \cite{b2}. This replication serves as a reference to fairly compare the proposed optimized models and assess the impact of pruning and knowledge distillation on IDS performance.
    
    In Figure \ref{fig2}, it is possible to observe a summary of the proposed system. It works like this: malicious data is purposely injected while the network is capturing benign data. These packets then pass through the Intrusion Detection System, which initially uses them to generate features and feeds them into a convolutional neural network. This network aims to classify whether the arrived packet is malicious or benign.
    It is possible to divide the construction of the proposed intrusion detection system into the following steps:

    \paragraph{Data pre-processing.} Initially, the processing and analysis of the data found in the dataset will be carried out. This process is important to understand the characteristics of the data and check if data processing is necessary before running the algorithm. Each AVTP packet in the dataset has a size of 438 bytes. The first 58 bytes of the packets contain patterns that vary from packet to packet, making it possible to distinguish between malicious and benign packets \cite{b1}. In the range 59-438, no significant pattern is found in the byte shift. In this way, only the first 58 bytes of the packets are used to generate features to be input into the model.

    \paragraph{Feature Generator.} After the data processing step, the feature generator is modeled. It will allow the CNN model to be able to find differences in the payload of packets after the insertion of malicious packets. {The Feature Generator output is used as input to the 2D-CNN model described in Section IV‑C (architecture from Jeong et al.~\cite{b1})}. This step defines a window value, where $w$ is an integer such that $w \geq 4$ {to capture temporal dependencies across consecutive packets; Jeong et al.~\cite{b1} found $w=44$ optimal.} The objective is to aggregate the traffic that arrived in windows of size $w$, producing a two-dimensional vector. The authors in \cite{b1} noted that a window where $w = 44$ brings the best accuracy values for the model to be built, so this window value will be used. A packet that arrives as input to the Feature Generator is represented by $X_{i}$, where $X_{i} = (X_{i,1}, X_{i,2}, ... , X_{i,(j-1)}, X_{i,58})$, $X_{i}$ is an integer between 0 and 256 and $j=58$, where j represents the j-th byte of the array. To generate the features, it is necessary to define a $\Delta X_i$, which represents the change of state of the i-th AVTP packet. $\Delta X_i$ is represented in Eq. \ref{eq1}, and code can be found in our GitHub repository\footnotemark.
    
\begin{figure*}[!ht]
    \centering
    \includegraphics[width=\textwidth,keepaspectratio]{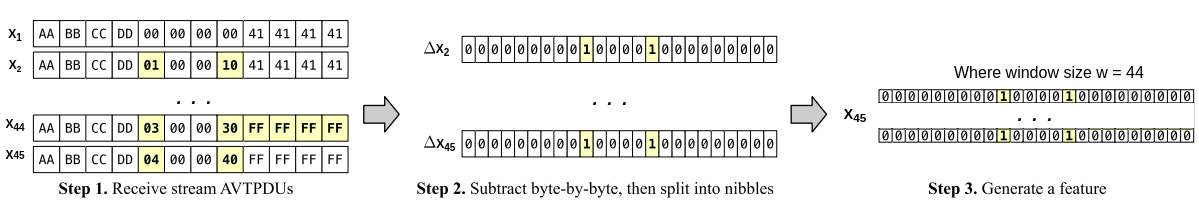}
    \caption{Feature Generator. Adapted from \cite{b1}.}
    \label{fig3}
    \end{figure*}

    In Fig \ref{fig3}, it is possible to observe the feature generation process. At the end of this step, a two-dimensional vector of size $w$ $\times$ $2j$ is generated. In this step, the labels of each generated feature are also created. Label 0 is assigned to those packets considered benign, and label 1 to packets considered injected. In this way, it is possible to train the proposed model using the data processed in this step.

    \begin{figure}[!h]
    \centering
    \includegraphics[width=1\textwidth]{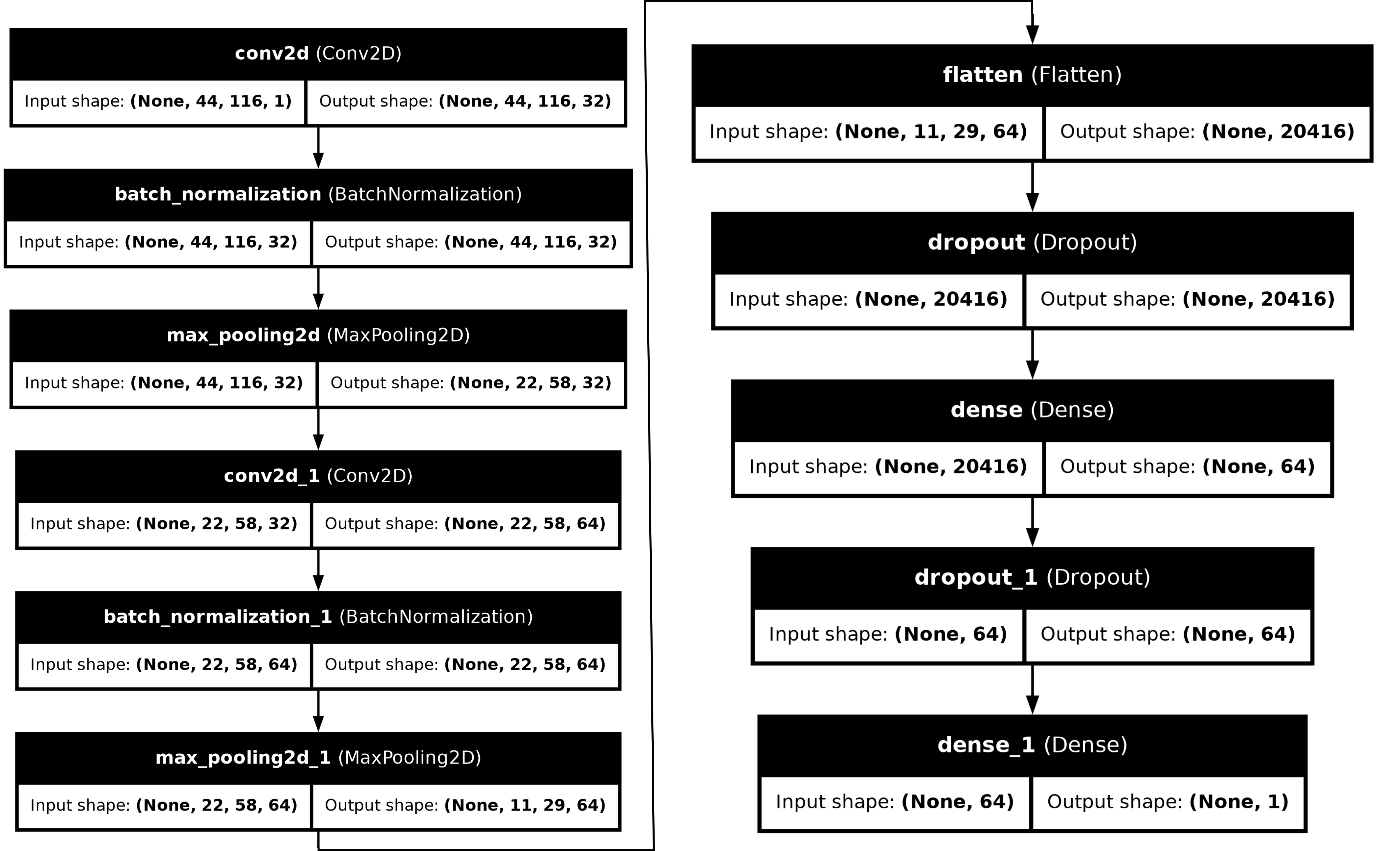}
    \caption{{Baseline 2D‑CNN: input 44×116×1, two conv sublayers (Conv2D + batch normalization + max pooling), two dense layers with dropout, and a sigmoid output for binary intrusion detection.}}
    \label{fig4}
    \end{figure}
    
    \begin{dmath}
    \Delta X_i \equiv (X_{i} - X_{i-1}) \; mod \; 2^8
    \equiv (X_{i,1}, X_{i,2}, ... , X_{i,(j-1)}, X_{i,58}) \; mod \; 2^8
    \equiv (\Delta X_{i,1}, \Delta X_{i,2}, ... , \Delta X_{i,(j-1)}, X_{i,58})
    \label{eq1}
    \end{dmath}
     
    \paragraph{Training a 2D-CNN model.} The model from \cite{b1} was implemented with the same network architecture and hyperparameters defined by the authors. TensorFlow v2.17.1 and Keras \cite{tensorflow_developers_2024_13989019} were used for the implementation in Python. The data $D_{indoors}$ has been used for training and validation, and the data $D_{driving}$ has been used as a test set. The network to be built comprises an input layer (which accepts $44 \times 116 \times 1$-sized data), two hidden sublayers, and two dense layers. Each hidden sublayer contains a convolution layer, a batch normalization layer, and a max-pooling layer. In layers conv2d, L2 regularization is used as kernel regularizer. Two dropout layers are added to prevent overfitting. ReLU (Rectified linear unit) is used as an activation function. The last layer uses sigmoid as the activation function and returns the value of $Y'_i$, which is a real value where $0 \leq Y'_i \leq 1$ and represents the percentage of the input being a malicious packet. Finally, Adam optimizer is used to train the model. Figure \ref{fig4} shows the structure of the 2D-CNN network used.  

    {\subsection{Experiments with Pruning and Quantization} To further optimize the 2D-CNN model for deployment in low-cost embedded systems, pruning and quantization techniques were explored. TensorFlow supports these methods \cite{tf_pruning_guide1, tf_pruning_guide2} and aims to reduce the model's size and improve inference times while maintaining its performance.}     
    {\paragraph{Pruning and Quantization Combined.} Pruning was applied to sparsify the weight matrices by reducing the magnitude of less significant weights, following a polynomial decay schedule. The pruning process started with 50\% sparsity and gradually increased to 90\% over training. After pruning, the model was converted to TensorFlow Lite (TFLite) format and quantized using the \texttt{TFLiteConverter}, which reduces the numerical precision of weights and activations (e.g., from 32-bit floats to 8-bit integers). This combination was designed to enable efficient deployment on resource-constrained devices, such as the Raspberry Pi 4.}
    {\paragraph{Pruning Only.} As an alternative, pruning alone was applied using the same polynomial decay schedule, targeting up to 90\% sparsity. This approach retained the original numerical precision of the model while reducing its size and sparsifying its structure. Both the pruned and pruned-and-quantized models were prepared for deployment using TensorFlow Lite.}

    Code implementations are available online\footnotemark[\value{footnote}].   
     
    {\subsection{Knowledge Distillation} Knowledge distillation is a training technique that leverages the predictions of a larger, well-trained model (the teacher) to train a smaller, more efficient model (the student). This approach allows the student model to mimic the teacher's behavior by minimizing a weighted combination of two losses: the standard cross-entropy loss on the true labels and the distillation loss on the soft labels produced by the teacher. The distillation loss is computed using the teacher's output probabilities, scaled by a temperature parameter \texttt{T}. Higher values of \texttt{T} produce softer probability distributions, enabling the student to learn richer information from the teacher's outputs \cite{knowledge_distillation}.}
    {In this work, the pre-trained 2D-CNN model described earlier was used as the teacher model. Two student models were designed:}
    {\paragraph{Student Model.} This model features two convolutional layers, each followed by batch normalization and max-pooling layers and two dense layers. The structure is designed to balance performance and efficiency, making it a smaller but effective alternative to the teacher model.}
    {\paragraph{Ultra-Light Student Model.} This highly compact model includes a single convolutional layer followed by a max-pooling layer and two dense layers. It significantly reduces the number of parameters, prioritizing deployment in environments with extreme resource constraints.}
    The distillation process was implemented using a custom framework inspired by the methodology in \cite{knowledge_distillation}. The training pipeline used the distillation loss function described above, with parameters \texttt{$\alpha$=0.1} and \texttt{temperature=3.0}. Additional strategies, such as learning rate reduction and early stopping, were employed to ensure efficient and robust training. Code implementations are available online\footnotemark[\value{footnote}].

    \footnotetext{The complete repository is available anonymously at: \href{https://anonymous.4open.science/r/DL-Based-IDS-2187}{anonymous.4open.science/r/DL-Based-IDS-2187}.}
    
\section{Experiments}

{The experiments were performed on four distinct hardware setups to evaluate the implemented models' performance and compatibility across various platforms. Each setup is detailed below:}

{The first machine used was the \textbf{Google\texttrademark{} Colab Pro}, equipped with a \textbf{NVIDIA\texttrademark{} Tesla P100 GPU}, 26GB of RAM, NVIDIA Driver version \textbf{460.32.03} and CUDA version \textbf{11.2}. This platform was chosen for its high-performance GPU and robust support for deep learning frameworks. Given that this GPU is a high-end model from 2016, it has a quite high thermal design power (TDP) of \textbf{250W}.}

{The second machine was an \textbf{NVIDIA Jetson Nano\texttrademark{}}, a compact and energy-efficient embedded system, with a TDP of \textbf{10W} for the entire device. Although the device has a GPU, all models were executed exclusively on the \textbf{Quad-core ARM Cortex-A57 CPU} with \textbf{4GB of RAM}. This decision was due to the Jetson Nano's GPU being compatible only with older versions of TensorFlow, which are incompatible with the models developed using TensorFlow 2.17.1. This setup reflects the challenges of deploying advanced deep learning models on edge devices with limited compatibility and constrained resources.}

{The third machine was a \textbf{Raspberry Pi 4 Model B} with \textbf{2GB of RAM} and a \textbf{Quad-core Cortex-A72 (ARM v8) 64-bit CPU}. This platform further explores deployment feasibility in extremely resource-limited environments, often encountered in IoT devices, on a low power budget of \textbf{10W}.}

{Finally, the fourth machine was a high-performance \textbf{desktop computer}, equipped with \textbf{48GB of RAM} and an \textbf{NVIDIA RTX 3060 GPU with 12GB of VRAM}. This setup serves as a benchmark for evaluating performance in a robust yet cost-effective desktop environment commonly used for deep learning research. It is also a much more efficient and newer device from 2021, with its GPU having a TDP of \textbf{170W}.}


{These diverse hardware configurations were selected to provide insight into the adaptability and efficiency of the models on platforms ranging from high-performance GPUs to resource-constrained edge devices. Table \ref{tab:hosts} summarizes these configurations.}

\begin{table}[!h]
    \caption{Devices used in experiments.}
    \centering
    \resizebox{\columnwidth}{!}{%
    \begin{tabular}{llc}
    \toprule
    \textbf{Host}            & \textbf{Processor}                  & \multicolumn{1}{l}{\textbf{Available RAM}} \\
    \toprule
    Google Collaboratory Pro & (GPU) NVIDIA Tesla P100             & 26GB                                       \\
    NVIDIA Jetson Nano       & (CPU) Quad-core ARM Cortex-A57    & 4GB                                        \\
    Raspberry Pi 4 Model B   & (CPU) Quad-core Cortex-A72           & 2GB                                        \\
    Desktop Computer         & (GPU) NVIDIA RTX 3060, 12GB VRAM    & 48GB                                       \\
    \bottomrule
    \end{tabular}%
    }
    \label{tab:hosts}
\end{table}

The data used to train the model were from the $D_{indoors}$ dataset, which contains 446,372 benign packets and 196,894 malicious packets. The $D_{driving}$ dataset was used for model testing and consists of 1,494,257 benign packets and 376,236 malicious packets.

\subsection{Evaluation metrics}
Performance metrics used were accuracy, precision, recall, F1-score, AUROC, and inference time per sample for real-time assessment.


\subsection{Model Train}

To train the 2D-CNN baseline model, stratified 5-fold cross-validation was used on the training data. At each step, 80\% and 20\% of the data are randomly selected as training and validation sets, respectively. Each input is used once as validation. There is the same proportion of benign/malicious examples in each cross-validation. As well as \cite{b1}, the batch size value is 64, and 30 epochs are used to train the model. At the end of the training, five models are obtained and evaluated using the test set, $D_{driving}$.

\section{Results}

Table I summarizes validation results (five-fold cross-validation) for the baseline 2D-CNN. F1-score (primary metric, due to class imbalance) improved from 0.9911–0.9947\cite{b1} to 0.9968–0.9993. This means that the models found can classify almost all packets correctly.

\begin{table}[!ht]
\label{t1}
\caption{Train results for 2D-CNN baseline.}
\centering
\begin{tabular}{rccccc}
\hline
\multicolumn{6}{c}{Classification results of five-fold cross-validation using dataset \textbf{Dindoors}}                                                                                                          \\ \hline
\multicolumn{1}{r|}{Fold}            & \multicolumn{1}{c|}{Accuracy}        & \multicolumn{1}{c|}{Precision}       & \multicolumn{1}{c|}{Recall}          & \multicolumn{1}{c|}{F1-score}        & AUC             \\ \hline
\multicolumn{1}{r|}{1}               & \multicolumn{1}{c|}{0.9991}          & \multicolumn{1}{c|}{0.9997}          & \multicolumn{1}{c|}{0.9990}          & \multicolumn{1}{c|}{0.9993}          & 0.9999          \\ \hline
\multicolumn{1}{r|}{2}               & \multicolumn{1}{c|}{0.9959}          & \multicolumn{1}{c|}{0.9944}          & \multicolumn{1}{c|}{0.9993}          & \multicolumn{1}{c|}{0.9968}          & 0.9967          \\ \hline
\multicolumn{1}{r|}{3}               & \multicolumn{1}{c|}{0.9986}          & \multicolumn{1}{c|}{0.9997}          & \multicolumn{1}{c|}{0.9984}          & \multicolumn{1}{c|}{0.9990}          & 0.9999          \\ \hline
\multicolumn{1}{r|}{4}               & \multicolumn{1}{c|}{0.9969}          & \multicolumn{1}{c|}{0.9989}          & \multicolumn{1}{c|}{0.9972}          & \multicolumn{1}{c|}{0.9981}          & 0.9995          \\ \hline
\multicolumn{1}{r|}{5}               & \multicolumn{1}{c|}{0.9987}          & \multicolumn{1}{c|}{0.9993}          & \multicolumn{1}{c|}{0.9991}          & \multicolumn{1}{c|}{0.9992}          & 0.9999          \\ \hline
\multicolumn{1}{r|}{Total (Average)} & \multicolumn{1}{c|}{\textbf{0.9978}} & \multicolumn{1}{c|}{\textbf{0.9984}} & \multicolumn{1}{c|}{\textbf{0.9986}} & \multicolumn{1}{c|}{\textbf{0.9985}} & \textbf{0.9992} \\ \hline
\end{tabular}
\end{table}

Table II shows the results for the test set on each of the five models created for the 2D-CNN baseline. All test-set metrics remained above 0.99, slightly lower than training results. Precision (0.9988–0.9994) and recall (0.9942–0.9962) notably surpassed previous values (precision: 0.9439–0.9821, F1-score: 0.9704–0.9885 in \cite{b1})

\begin{table}[!ht]
\centering
\label{t2}
\caption{Test results for 2D-CNN baseline.}
\begin{tabular}{lccccc}
\hline
\multicolumn{6}{c}{Classification results using dataset \textbf{Ddriving}}                                                                                                                                                 \\ \hline
\multicolumn{1}{r|}{Model}           & \multicolumn{1}{c|}{Accuracy}        & \multicolumn{1}{c|}{Precision}       & \multicolumn{1}{c|}{Recall}          & \multicolumn{1}{c|}{F1-score}        & AUC             \\ \hline
\multicolumn{1}{r|}{1}               & \multicolumn{1}{c|}{0.9985}          & \multicolumn{1}{c|}{0.9989}          & \multicolumn{1}{c|}{0.9959}          & \multicolumn{1}{c|}{0.9974}          & 0.9993          \\ \hline
\multicolumn{1}{r|}{2}               & \multicolumn{1}{c|}{0.9984}          & \multicolumn{1}{c|}{0.9988}          & \multicolumn{1}{c|}{0.9955}          & \multicolumn{1}{c|}{0.9972}          & 0.9995          \\ \hline
\multicolumn{1}{r|}{3}               & \multicolumn{1}{c|}{0.9987}          & \multicolumn{1}{c|}{0.9993}          & \multicolumn{1}{c|}{0.9962}          & \multicolumn{1}{c|}{0.9978}          & 0.9994          \\ \hline
\multicolumn{1}{r|}{4}               & \multicolumn{1}{c|}{0.9985}          & \multicolumn{1}{c|}{0.9993}          & \multicolumn{1}{c|}{0.9952}          & \multicolumn{1}{c|}{0.9972}          & 0.9992          \\ \hline
\multicolumn{1}{r|}{5}               & \multicolumn{1}{c|}{0.9982}          & \multicolumn{1}{c|}{0.9994}          & \multicolumn{1}{c|}{0.9942}          & \multicolumn{1}{c|}{0.9968}          & 0.9990          \\ \hline
\multicolumn{1}{r|}{Total (Average)} & \multicolumn{1}{c|}{\textbf{0.9984}} & \multicolumn{1}{c|}{\textbf{0.9991}} & \multicolumn{1}{c|}{\textbf{0.9954}} & \multicolumn{1}{c|}{\textbf{0.9973}} & \textbf{0.9993} \\ \hline
\end{tabular}
\end{table}

Since we already have the baseline model results, we can compare them with the other models created. Table \ref{tab:modelResults} summarizes the performance (accuracy, precision, recall, F1, AUROC) of optimized models compared to the baseline 2D-CNN model (Full Model).

The metrics for the Full Model represent the averages obtained from a comprehensive 5-fold cross-validation process, ensuring the robustness and reliability of the reported values. In contrast, the other models, including the Pruning Model, Pruning and Quantization, Distilling Model, and Ultra Light Distilling Model, were evaluated using a single fold, and their results reflect the performance from that specific test run. {The five evaluated models differ as follows:}
{\begin{itemize}
  \item \textbf{Full Model:} Baseline 2D‑CNN architecture (no optimizations).
  \item \textbf{Pruning Model:} Baseline pruned to 90\% sparsity via a polynomial decay schedule.
  \item \textbf{Pruning + Quantization:} Pruned model further quantized to 8‑bit integer weights/activations in TFLite.
  \item \textbf{Distilling Model:} Smaller student network (two conv layers + two dense layers) trained via knowledge distillation (\(\alpha=0.1\), \(T=3.0\)).
  \item \textbf{Ultra‑Light Distilling Model:} Highly compact student network (one conv layer + two dense layers) trained via the same distillation pipeline.
\end{itemize}}

\begin{table*}[!h]
    \caption{Classification results for different models on the test dataset.}
    \label{tab:modelResults}
    \centering
    \begin{adjustbox}{max width=\textwidth}
    \begin{tabular}{lccccc}
    \toprule
    \textbf{Model} & \textbf{Accuracy} & \textbf{Precision} & \textbf{Recall} & \textbf{F1-score} & \textbf{AUROC} \\
    \toprule
    \textbf{Full Model}              & 0.9984          & 0.9991          & 0.9954          & 0.9973          & 0.9993 \\
    \textbf{Pruning Model}           & 0.9980          & 0.9989          & 0.9948          & 0.9963          & 0.9990 \\
    \textbf{Pruning and Quantization} & 0.9980         & 0.9989          & 0.9948          & 0.9963          & 0.9990 \\
    \textbf{Distilling Model}        & 0.9981          & 0.9990          & 0.9949          & 0.9972          & 0.9990 \\
    \textbf{Ultra Light Distilling Model} & 0.9865          & 0.9953          & 0.9562          & 0.9754          & 0.9890 \\
    \bottomrule
    \end{tabular}
    \end{adjustbox}
\end{table*}

{The results in Table \ref{tab:modelResults} reveal interesting trends across the evaluated models, particularly regarding the performance of the Pruning Model, Pruning and Quantization Model, and Distilling Model. These models exhibit metrics that are strikingly close to the Full Model, with only minor reductions in accuracy, precision, recall, and F1-score. However, the Ultra Light Distilling Model demonstrates a more pronounced drop in performance.}

\begin{figure}[htbp]
  \centering
  \begin{subfigure}[t]{0.48\textwidth}
    \centering
    \includegraphics[
      width=\linewidth,
      height=0.5\textheight,
      keepaspectratio
    ]{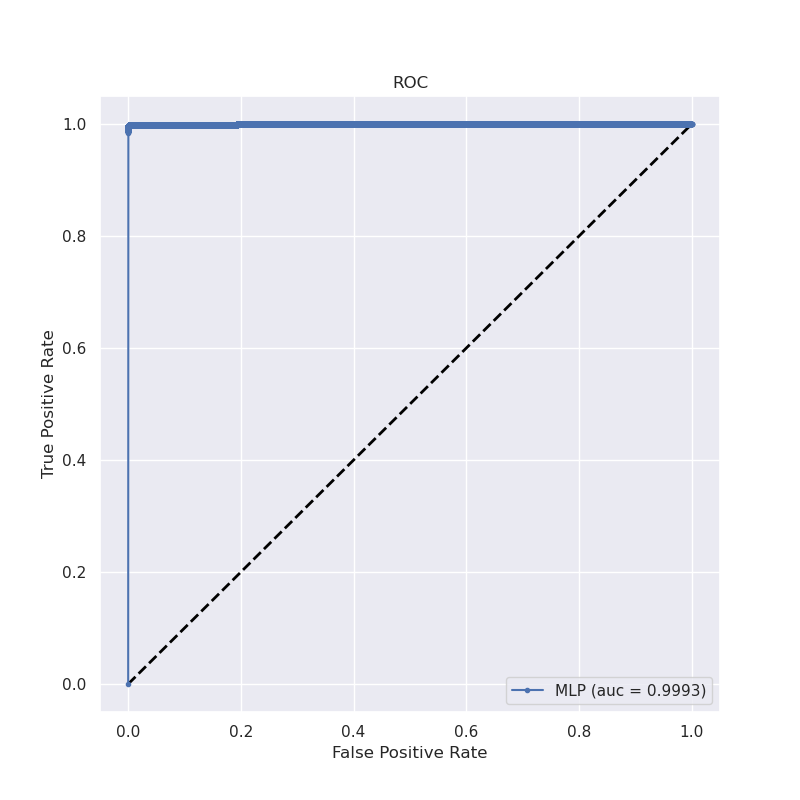}
    \caption{Full Model}
    \label{fig:roc_full_model}
  \end{subfigure}\hfill
  \begin{subfigure}[t]{0.48\textwidth}
    \centering
    \includegraphics[
      width=\linewidth,
      height=0.5\textheight,
      keepaspectratio
    ]{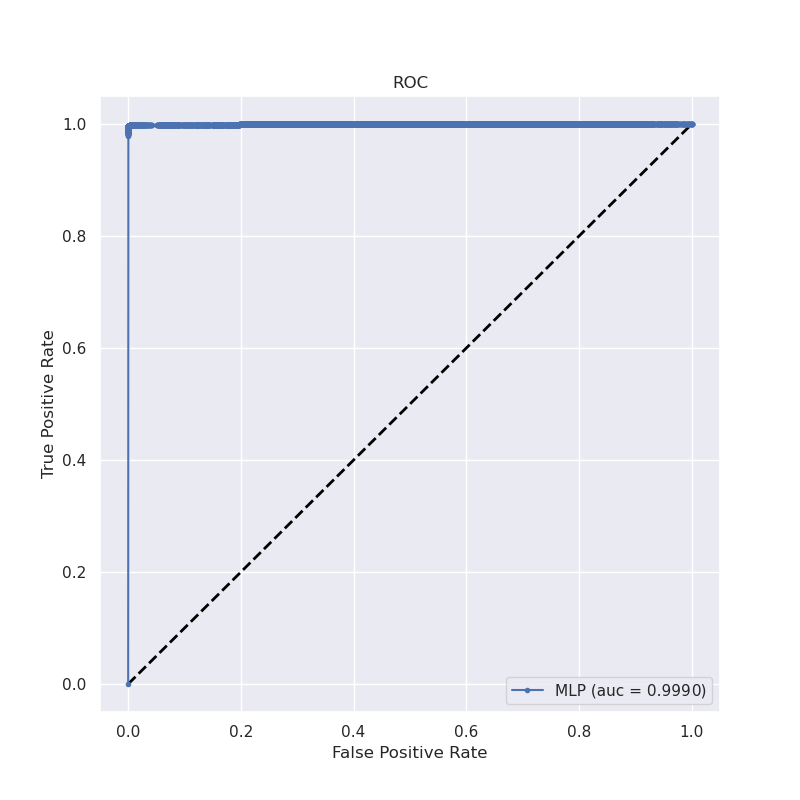}
    \caption{Pruned Model}
    \label{fig:roc_pruned_model}
  \end{subfigure}

  \vspace{0.1em}

  \begin{subfigure}[t]{0.48\textwidth}
    \centering
    \includegraphics[
      width=\linewidth,
      height=0.5\textheight,
      keepaspectratio
    ]{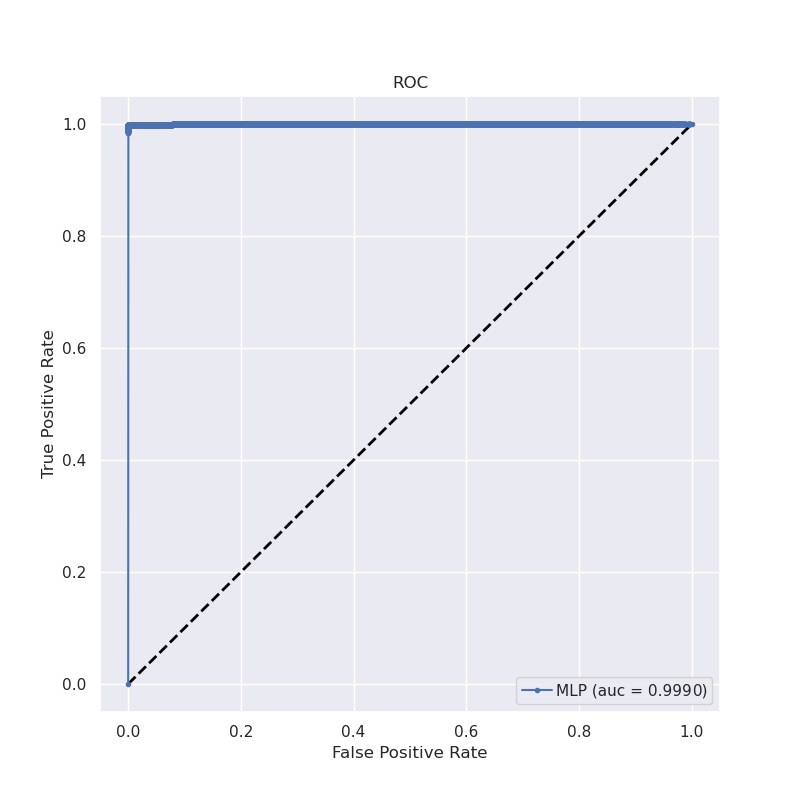}
    \caption{Distilling Model}
    \label{fig:roc_student_full_model}
  \end{subfigure}\hfill
  \begin{subfigure}[t]{0.48\textwidth}
    \centering
    \includegraphics[
      width=\linewidth,
      height=0.5\textheight,
      keepaspectratio
    ]{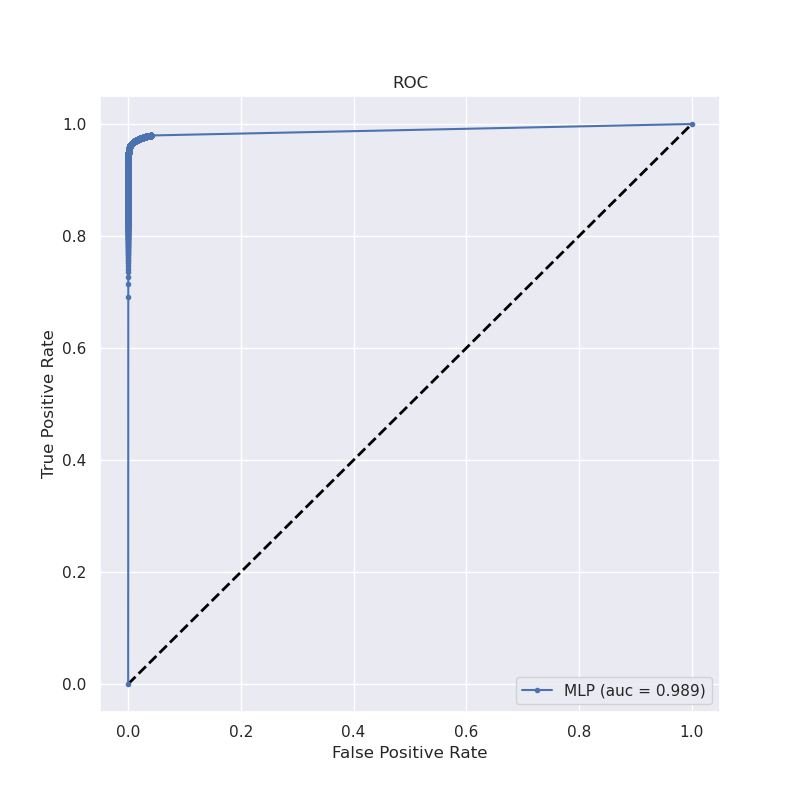}
    \caption{Ultra-Light Distilling Model}
    \label{fig:roc_ultra_light_student_model}
  \end{subfigure}

  \caption{ROC Curves for different IDS models (Test set).}
  \label{fig:roc_curves}
\end{figure}

Figure \ref{fig:roc_curves} presents the Receiver Operating Characteristic (ROC) curves for the evaluated Intrusion Detection System (IDS) models in the test set. The curves include the Full Model (\ref{fig:roc_full_model}), the Pruned Model (\ref{fig:roc_pruned_model}), the Distilling Model (\ref{fig:roc_student_full_model}), and the Ultra-Light Distilling Model (\ref{fig:roc_ultra_light_student_model}). These visualizations provide a comparative analysis of each model's performance by illustrating the trade-off between the true positive rate (TPR) and the false positive rate (FPR).

\begin{table*}[!h]
    \caption{Average inference time per sample for the trained models.}
    \label{tab:inference_results}
    \centering
    \begin{adjustbox}{max width=\textwidth}
    \begin{tabular}{lccccc}
    \toprule
    \textbf{Host} & 
    \textbf{\begin{tabular}[c]{@{}c@{}}Full Model\\ (\textmu s/sample)\end{tabular}} & 
    \textbf{\begin{tabular}[c]{@{}c@{}}Pruning Model\\ (\textmu s/sample)\end{tabular}} & 
    \textbf{\begin{tabular}[c]{@{}c@{}}Pruning and\\ Quantization\\ (\textmu s/sample)\end{tabular}} & 
    \textbf{\begin{tabular}[c]{@{}c@{}}Distilling Model\\ (\textmu s/sample)\end{tabular}} & 
    \textbf{\begin{tabular}[c]{@{}c@{}}Ultra Light Distilling Model\\ (\textmu s/sample)\end{tabular}} \\
    \toprule
    Google Collaboratory Pro & 60.44 & --    & --    & --    & -- \\
    Desktop with RTX 3060 12GB & 78.36 & 79.88    & 78.69    & 62.45 & 47.83 \\
    NVIDIA Jetson Nano CPU     & 21301   & 20204    & 20142    & 2904  & 727.51   \\
    Raspberry Pi 4 Model B     & 17259 & 17043 & 16550 & 3006  & 849.78     \\
    \bottomrule
    \end{tabular}
    \end{adjustbox}
\end{table*}

\begin{table*}[!h]
    \caption{Energy efficiency of distilled models across different devices (excluding Google Collaboratory Pro).}
    \label{tab:efficiency_distilled_models}
    \centering
    \begin{adjustbox}{max width=\textwidth}
    \begin{tabular}{lccccc}
    \toprule
    \textbf{Host} & 
    \textbf{TDP (W)} & 
    \textbf{\begin{tabular}[c]{@{}c@{}}Throughput\\ Distilled\\ (samples/sec)\end{tabular}} & 
    \textbf{\begin{tabular}[c]{@{}c@{}}Throughput\\ Ultra Light\\ (samples/sec)\end{tabular}} & 
    \textbf{\begin{tabular}[c]{@{}c@{}}Efficiency\\ Distilled\\ (samples/sec/W)\end{tabular}} & 
    \textbf{\begin{tabular}[c]{@{}c@{}}Efficiency\\ Ultra Light\\ (samples/sec/W)\end{tabular}} \\
    \toprule
    Desktop with RTX 3060 12GB & 170 & 16,012.81 & 20,907.38 & 94.19 & 122.98 \\
    NVIDIA Jetson Nano CPU     & 10  & 344.35 & 1,374.55 & 34.44 & 137.46 \\
    Raspberry Pi 4 Model B     & 10  & 332.67 & 1,176.78 & 33.27 & 117.68 \\
    \bottomrule
    \end{tabular}
    \end{adjustbox}
\end{table*}

{Table \ref{tab:inference_results} shows average inference times across all models and hardware setups.. The results demonstrate the performance of these models on devices ranging from high-performance GPUs to resource-constrained embedded systems.}

{Using the \textbf{Google Colab Pro} with an NVIDIA Tesla P100 GPU, the average inference time for the \textbf{full model} was \textbf{60.44 µs/sample}. This value remains well below the real-time detection threshold of \textbf{1000 µs/sample}, showcasing the platform’s suitability for time-sensitive applications.}

{On the \textbf{Desktop Computer} equipped with an NVIDIA RTX 3060 GPU, the full model achieved an inference time of \textbf{44.25 µs/sample}. Optimized versions of the model delivered even better results, with the \textbf{distilled model} achieving an average inference time of \textbf{31.20 µs/sample}, while the \textbf{ultra-light distilled model} reduced the time further to \textbf{29.98 µs/sample}. These results underscore the RTX 3060’s capability to deliver highly efficient inference times, making it well-suited for real-time and time-critical applications.}

The \textbf{NVIDIA Jetson Nano}, a low-power edge computing device, recorded an inference time of \textbf{21301 µs/sample} for the full model. The \textbf{distilled model} achieved \textbf{2904 µs/sample}, and the \textbf{ultra-light distilled model} reached \textbf{727.51 µs/sample}, making it feasible for deployment in constrained environments with appropriate optimizations.

{The \textbf{Raspberry Pi 4 Model B}, characterized by its limited computational resources, recorded an inference time of \textbf{17259 µs/sample} for the full model, far exceeding the real-time threshold. However, pruning techniques did not lead to significant improvements, as the \textbf{pruned model} recorded a similar inference time of \textbf{17043 µs/sample}, and the \textbf{pruned and quantized model} only marginally improved performance to \textbf{16550 µs/sample}. In contrast, distillation techniques demonstrated far better results, with the \textbf{distilled model} achieving \textbf{3006 µs/sample}. Notably, the \textbf{ultra-light distilled model} was the only variant capable of meeting the real-time threshold, with an inference time of \textbf{849.78 µs/sample}, highlighting that distilling can enable viable deployments on resource-constrained hardware.}

Table \ref{tab:efficiency_distilled_models} presents the energy efficiency of distilled and ultra-light distilled models. The RTX 3060 achieved the highest throughput, with up to 20,907.38 samples/sec and an energy efficiency of 122.98 samples/sec/W for the ultra-light model. The Jetson Nano, while lower in throughput, achieved the highest energy efficiency of 137.46 samples/sec/W. The Raspberry Pi 4 showed limited throughput but maintained reasonable energy efficiency at 117.68 samples/sec/W for the ultra-light model.

\section{Discussion of Results}

{The results from Table \ref{tab:modelResults} and the corresponding inference times provide valuable insights into the trade-offs between model performance, complexity, and deployment feasibility. While pruning and distillation techniques showed promise in reducing model sizes and computational costs, their impacts on inference times and predictive performance reveal distinct patterns worth discussing.}

\subsection*{Performance Proximity Across Models}
{The Pruning Model, Pruning and Quantization Model, and Distilling Model closely matched the baseline model's performance, indicating that these optimization techniques effectively preserved the core functionality of the original model. Pruning, in particular, excluded weights and connections that had minimal contributions to the decision-making process, while quantization compressed the model without introducing significant errors. Similarly, the distillation process successfully transferred the knowledge from the Full Model to a smaller architecture, ensuring that critical patterns were retained.}

{However, despite these similarities in performance, the reductions in inference times achieved by these techniques were insufficient to meet real-time constraints on low-cost embedded devices. This suggests that while the pruning and distillation techniques compress the model, the computational bottlenecks inherent in such hardware platforms may still impose significant limitations. The preserved decision-making complexity of these models, which benefits predictive performance, likely contributes to their inability to achieve the drastic inference speed gains needed for specific applications.}

\subsection*{Ultra Light Distilling Model: A Feasible Alternative}
{Only the Ultra-Light model achieved real-time inference on constrained devices, albeit with minor accuracy (\textbf{0.9865}) and recall (\textbf{0.9562}) reductions due to aggressive simplification.}

{Despite this reduction in performance, the Ultra Light Distilling Model maintained a high AUROC (\textbf{0.9890}), indicating it still performs well overall. This balance of efficiency and predictive power makes it a strong candidate for applications where computational resources are limited but near-real-time predictions are required.}

\subsection*{Energy Efficiency Considerations for Deployment}
\textcolor{black}{While inference speed and model accuracy are key deployment considerations, energy efficiency plays a crucial role in real-world applications. The results in Table \ref{tab:efficiency_distilled_models} show that, despite achieving the lowest inference times, the RTX 3060 has a performance-per-watt efficiency that's on par with the embedded platforms like the Jetson Nano and Raspberry Pi 4 on the \textit{Ultra Light Distillation} model. However, it’s important to note that the RTX 3060’s TDP only accounts for the GPU itself. In contrast, the embedded devices include the entire system’s power consumption, making their efficiency even more favorable for constrained environments. This highlights a trade-off: while bigger GPUs offer raw performance, they have a significant energy overhead, making them less practical for always-on, low-power applications like automotive ECUs or edge AI deployments.}

\textcolor{black}{This work focused on minimizing inference time on embedded devices rather than optimizing energy efficiency, leaving room for future improvements. One potential enhancement is leveraging the Jetson Nano’s GPU acceleration, which could significantly improve throughput without increasing power consumption. Similarly, specialized ARM-optimized libraries for the Raspberry Pi could further boost efficiency. Future research should explore these optimizations to strike a better balance between inference speed, energy consumption, and real-world feasibility, ensuring lightweight models remain viable for deployment in resource-constrained environments.}

\subsection*{When to Choose Lightweight Models}
{Lightweight models, like the Ultra Light Distilling Model, become particularly advantageous in scenarios where rapid response times are essential for maintaining the safety and security of the vehicle's network. The slight trade-off in accuracy and recall may be acceptable in favor of ensuring real-time threat detection and mitigation.}

{The ability to quickly process and classify network data is crucial for intrusion detection, as delays in identifying malicious activity could result in compromised safety-critical functions. Moreover, the cost constraints in automotive systems often make relying on GPUs or other high-performance hardware impractical, necessitating solutions that can operate effectively on low-cost embedded devices. With its ability to achieve inference times within real-time thresholds, the Ultra Light Distilling Model offers a feasible solution for deployment on resource-constrained automotive hardware, such as ECUs or other embedded systems, ensuring both affordability and efficiency.}

{Additionally, this model can be deployed as a fallback or auxiliary system when the vehicle's more powerful processing units are overloaded with other critical tasks. By offloading intrusion detection to a lightweight and efficient model, the system can remain continuously active, ensuring uninterrupted monitoring of the network and maintaining security without compromising the performance of other essential vehicle functions.}

\section{Conclusions}

{The paper explored the feasibility of applying optimization techniques, such as \textit{pruning} and \textit{distillation} to optimize DL-based IDS models for real-time automotive Ethernet intrusion detection on low-cost platforms. While \textit{distillation} demonstrated significant improvements in computational efficiency and inference time, \textit{pruning} alone was less effective in achieving the drastic reductions needed to meet real-time constraints on these resource-limited devices. The \textit{Ultra Light Distillation} model proved particularly viable, achieving inference times as low as 727 $\mu$s on a Raspberry Pi 4 while maintaining an AUC-ROC of 0.9890. This balance of performance and efficiency makes it a practical solution for real-time detection, especially in scenarios where rapid response is critical for ensuring the safety of the vehicle network. However, this came with a slight reduction in precision and \textit{recall}, a trade-off that may be acceptable in exchange for computational efficiency in constrained environments.}

{Limitations include the incompatibility of devices such as the Jetson Nano with newer versions of TensorFlow and the challenges of running full models on resource-constrained hardware. The following steps suggest exploring other fast inference techniques, such as \textit{early exit}; validating the models in real vehicular systems; investigating their robustness against adversarial attacks, {and examining on‑device training and continual learning strategies to continuously adapt to novel attack patterns.}

\FloatBarrier

\bibliographystyle{IEEEtran}
\bibliography{references}

\begin{thebibliography}{10}
\providecommand{\url}[1]{#1}
\csname url@samestyle\endcsname
\providecommand{\newblock}{\relax}
\providecommand{\bibinfo}[2]{#2}
\providecommand{\BIBentrySTDinterwordspacing}{\spaceskip=0pt\relax}
\providecommand{\BIBentryALTinterwordstretchfactor}{4}
\providecommand{\BIBentryALTinterwordspacing}{\spaceskip=\fontdimen2\font plus
\BIBentryALTinterwordstretchfactor\fontdimen3\font minus \fontdimen4\font\relax}
\providecommand{\BIBforeignlanguage}[2]{{%
\expandafter\ifx\csname l@#1\endcsname\relax
\typeout{** WARNING: IEEEtran.bst: No hyphenation pattern has been}%
\typeout{** loaded for the language `#1'. Using the pattern for}%
\typeout{** the default language instead.}%
\else
\language=\csname l@#1\endcsname
\fi
#2}}
\providecommand{\BIBdecl}{\relax}
\BIBdecl

\bibitem{porter2018100base}
D.~Porter, ``{100BASE-T1 Ethernet: the evolution of automotive networking},'' \emph{Texas Instruments, Techn. Ber}, 2018.

\bibitem{matheus2021automotive}
K.~Matheus and T.~Königseder, \emph{{Automotive Ethernet}}, 3rd~ed.\hskip 1em plus 0.5em minus 0.4em\relax Cambridge University Press, 2021.

\bibitem{8053478}
J.~Liu, S.~Zhang, W.~Sun, and Y.~Shi, ``{In-Vehicle Network Attacks and Countermeasures: Challenges and Future Directions},'' \emph{IEEE Network}, vol.~31, no.~5, pp. 50--58, 2017.

\bibitem{8688625}
W.~Wu, R.~Li, G.~Xie, J.~An, Y.~Bai, J.~Zhou, and K.~Li, ``{A Survey of Intrusion Detection for In-Vehicle Networks},'' \emph{IEEE Transactions on Intelligent Transportation Systems}, vol.~21, no.~3, pp. 919--933, 2020.

\bibitem{b2}
\BIBentryALTinterwordspacing
S.~Jeong, B.~Jeon, B.~Chung, and H.~K. Kim, ``Automotive ethernet intrusion dataset,'' 2021. [Online]. Available: \url{https://dx.doi.org/10.21227/1yr3-q009}
\BIBentrySTDinterwordspacing

\bibitem{b1}
------, ``Convolutional neural network-based intrusion detection system for avtp streams in automotive ethernet-based networks,'' \emph{Vehicular Communications}, vol.~29, p. 100338, 2021.

\bibitem{hinton2015distillingknowledgeneuralnetwork}
\BIBentryALTinterwordspacing
G.~Hinton, O.~Vinyals, and J.~Dean, ``Distilling the knowledge in a neural network,'' 2015. [Online]. Available: \url{https://arxiv.org/abs/1503.02531}
\BIBentrySTDinterwordspacing

\bibitem{b7}
S.~Mehta and M.~Rastegari, ``Mobilevit: Light-weight, general-purpose, and mobile-friendly vision transformer,'' \emph{arXiv preprint arXiv:2110.02178}, 2021.

\bibitem{mobilenet}
A.~G. Howard \emph{et~al.}, ``Mobilenets: Efficient convolutional neural networks for mobile vision applications,'' \emph{arXiv preprint arXiv:1704.04861}, 2017.

\bibitem{tf_pruning_guide1}
\BIBentryALTinterwordspacing
TensorFlow, ``Pruning with keras,'' 2024, tensorFlow Model Optimization Guide. [Online]. Available: \url{https://www.tensorflow.org/model\_optimization/guide/pruning/pruning\_with\_keras}
\BIBentrySTDinterwordspacing

\bibitem{tf_pruning_guide2}
\BIBentryALTinterwordspacing
------, ``Comprehensive guide for pruning,'' 2024, tensorFlow Model Optimization Guide. [Online]. Available: \url{https://www.tensorflow.org/model\_optimization/guide/pruning/comprehensive\_guide}
\BIBentrySTDinterwordspacing

\bibitem{tensorflowL2}
\BIBentryALTinterwordspacing
T.~C. Team, ``tf.keras.regularizers.l2,'' 2024. [Online]. Available: \url{https://www.tensorflow.org/api\_docs/python/tf/keras/regularizers/L2}
\BIBentrySTDinterwordspacing

\bibitem{b8}
P.~R. Carmo, P.~F. de~Araujo-Filho, D.~R. Campelo, E.~Freitas, A.~T. de~Oliveira~Filho, and D.~F. Sadok, ``Machine learning-based intrusion detection system for automotive ethernet: Detecting cyber-attacks with a low-cost platform,'' in \emph{Anais do XL Simp{\'o}sio Brasileiro de Redes de Computadores e Sistemas Distribu{\'\i}dos}.\hskip 1em plus 0.5em minus 0.4em\relax SBC, 2022, pp. 196--209.

\bibitem{tensorflow_developers_2024_13989019}
\BIBentryALTinterwordspacing
T.~Developers, ``Tensorflow,'' Oct. 2024. [Online]. Available: \url{https://doi.org/10.5281/zenodo.13989019}
\BIBentrySTDinterwordspacing

\bibitem{knowledge_distillation}
K.~Team, ``Knowledge distillation example,'' \url{https://keras.io/examples/vision/knowledge_distillation/}, 2023, accessed: 2024-12-10.

\end{thebibliography}

\end{document}